\pdfoutput=1

\documentclass[11pt]{article}
\usepackage{emnlp2023}

\usepackage{booktabs} 
\usepackage{times}
\usepackage{latexsym}
\usepackage{color, colortbl}
\usepackage{tabularx}
\usepackage{array}
\usepackage{amsmath, amssymb}
\usepackage{xcolor}
\usepackage{pifont} 
\usepackage{adjustbox}
\usepackage{natbib}
\definecolor{darkgreen}{RGB}{0,150,0}  
\definecolor{darkred}{RGB}{200,0,0}

\definecolor{bestgreen}{RGB}{0, 128, 0}
\definecolor{worstred}{RGB}{180, 0, 0}

\usepackage[T1]{fontenc}
\usepackage[utf8]{inputenc}
\usepackage{microtype}
\usepackage{inconsolata}
\usepackage{graphicx}
\usepackage{amsmath} 
\usepackage{enumitem}
\usepackage{titlesec}
\usepackage[most]{tcolorbox}
\usepackage{xcolor}
\usepackage{lipsum} 
\usepackage{multirow}
\usepackage{amssymb}
\usepackage{listings}
\usepackage{xcolor}
\usepackage[framemethod=default]{mdframed}
\usepackage{caption}

\newmdenv[
  skipabove=0.5em,
  skipbelow=0.5em,
  innertopmargin=4pt,
  innerbottommargin=4pt,
  innerleftmargin=6pt,
  innerrightmargin=6pt,
  linecolor=black!15,
  linewidth=0.5pt,
  backgroundcolor=gray!3
]{promptblock}

\lstdefinelanguage{json}{
    morestring=[b]",
    morecomment=[l]{//},
    stringstyle=\color{black},
    commentstyle=\color{gray},
    basicstyle=\ttfamily\small,
    keywordstyle=\color{blue},
    morekeywords={true,false,null}
}

\newcommand{\up}{\raisebox{0pt}{\scriptsize$\uparrow$}}
\newcommand{\down}{\raisebox{0pt}{\scriptsize$\downarrow$}}
\newcommand{\same}{\raisebox{0pt}{\scriptsize$\approx$}}

\definecolor{lightblue}{RGB}{230, 245, 255}

\titlespacing*{\paragraph}{0pt}{0.3ex plus 0.2ex minus 0.2ex}{1em}
\usepackage{xspace}  
\newcommand{\name}{\textsc{ClimateViz}\xspace}

\title{
\name: A Benchmark for Statistical Reasoning and \\ Fact Verification on Scientific Charts}

\author{
  Ruiran Su$^{1}$,
  Jiasheng Si$^{2}$,
  Zhijiang Guo$^{3, 4}$,
  Janet B. Pierrehumbert$^{1}$ \\
  $^{1}$University of Oxford \\
  $^{2}$Qilu University of Technology(Shandong Academy of Sciences) \\
  $^{3}$Hong Kong University of Science and Technology \\
  $^{4}$Hong Kong University of Science and Technology (Guangzhou) \\
  \texttt{ruiran.su@trinity.ox.ac.uk}, 
  \texttt{jiashengsi@qlu.edu.cn}, \\
  \texttt{zhijiangguo@hkust-gz.edu.cn}, 
  \texttt{janet.pierrehumbert@oerc.ox.ac.uk}
}

\begin{document}
\maketitle
\begin{abstract}
Scientific fact-checking has largely focused on textual and tabular sources, neglecting scientific charts—a primary medium for conveying quantitative evidence and supporting statistical reasoning in research communication. We introduce \textsc{ClimateViz}, the first large-scale benchmark for scientific fact-checking grounded in real-world, expert-curated scientific charts. \textsc{ClimateViz} comprises 49,862 claims paired with 2,896 visualizations, each labeled as support, refute, or not enough information. To enable interpretable verification, each instance includes structured knowledge graph explanations that capture statistical patterns, temporal trends, spatial comparisons, and causal relations. We conduct a comprehensive evaluation of state-of-the-art multimodal large language models, including proprietary and open-source systems, under zero-shot and few-shot settings. Our results show that current models struggle to perform fact-checking when statistical reasoning over charts is required: even the best-performing systems, such as Gemini 2.5 and InternVL 2.5, achieve only 76.2–77.8\% accuracy in label-only output settings, which is far below human performance (89.3\% and 92.7\%). While few-shot prompting yields limited improvements, explanation-augmented outputs significantly enhance performance in some closed-source models, notably o3 and Gemini 2.5. We released our dataset and code alongside the paper.\footnote{\url{https://github.com/Albasu120491/ClimateViz}}

\end{abstract}

\section{Introduction}

Scientific fact-checking—the task of assessing the validity of scientific claims through cross-referencing with established literature, empirical observations, or experimental data \citep{wadden-2020-scifact,vladika2023scientificfactcheckingsurveyresources}—is essential for maintaining the integrity of research findings, combating misinformation, and preserving public confidence in scientific discourse \citep{wadden2022scifactopenopendomainscientificclaim}. However, the rapid accumulation of scholarly findings and the increasing demand for domain-specific expertise often exceed the capacity of manual verification, making scientific fact-checking a critical focus in the NLP community.

\begin{figure*}[t]
  \centering
  \includegraphics[width=\textwidth]{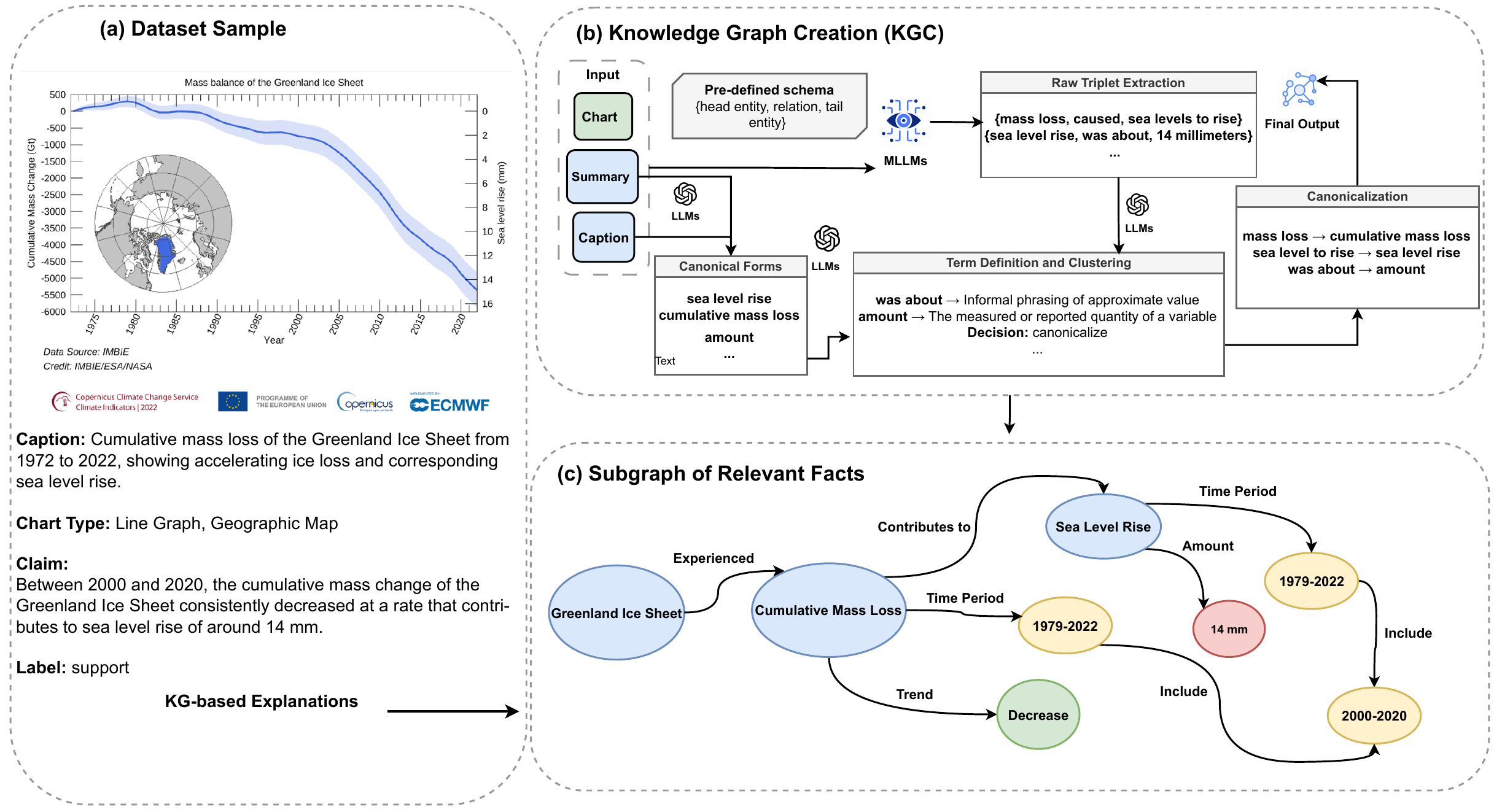}
  \caption{(a) A sample from \name showing a scientific chart, caption, claim, and its label.
(b) Knowledge graph creation pipeline: raw triplets are extracted by a multimodal LLM and canonicalized.
(c) Subgraph of relevant facts representing structured reasoning from the chart to the claim.}
  \label{fig:sample}
\end{figure*}

Significant progress has been made with the development of benchmarks such as SciFact \citep{wadden-2020-scifact}, SciFact-Open \citep{wadden2022scifactopenopendomainscientificclaim}, and SciTab \citep{lu-2023-scitab}. Despite these advances, existing resources exhibit critical limitations in scope. Specifically, prior benchmarks predominantly focus on verifying scientific claims against \textit{textual} \citep{wadden-2020-scifact,diggelmann2021climatefeverdatasetverificationrealworld,healthver-2021-evidence-based,saakyan2021covidfact} or \textit{tabular} \citep{2022-covert,lu-2023-scitab} evidence, particularly from literature abstracts or tables. These claims are typically validated using \textit{semantic} or \textit{structural} logical reasoning between claims and corresponding evidence. In contrast, real-world scientific findings often involve claims that are intrinsically tied to quantitative data. In such contexts, charts serve as both visual and statistical representations, summarizing complex numerical information, revealing trends, supporting quantitative reasoning, and effectively communicating scientific insights \citep{huang2024pixelsinsightssurveyautomatic}. However, chart-based verification is largely absent from prior fact-checking benchmarks—despite requiring explicit \textit{chart understanding} and \textit{statistical reasoning} over visualized data, beyond what semantic or structural inference alone can handle \citep{AkhtarSGCS023, akhtar2024chartcheckexplainablefactcheckingrealworld}.

More specifically, scientific claims often involve not only raw data observations but also interpretations of patterns, anomalies, correlations, and statistical aggregation. For example, the claim in Figure~\ref{fig:sample} implies a statistical relationship between \textit{ice sheet mass loss} and \textit{sea-level rise}. Verifying such information is inherently challenging and requires: 
(i) \textit{visual understanding} of charts with non-uniform temporal granularity (e.g., dual-axis time series), spatial dimensions (e.g., geographic insets), and uncertainty bands; (ii) \textit{nuanced statistical reasoning} to interpret temporal trends, quantify cumulative change, and relate ice mass loss to sea-level rise; and (iii) \textit{cross-modal reasoning} to establish logical coherence between the linguistic embedding of scientific claims and the perceptual features of the chart. This example illustrates the depth of reasoning needed to understand complex scientific findings, yet the absence of benchmarks that require such statistical reasoning severely limits the evaluation of models designed for general scientific understanding.

In this paper, we introduce \name, a novel dataset sourced from reputable climate institutions (e.g., the National Oceanic and Atmospheric Administration\footnote{\url{https://www.noaa.gov/}} and the UK Met Office\footnote{\url{https://www.metoffice.gov.uk/}}), designed to advance scientific fact-checking with a focus on statistical reasoning over charts. \name comprises 49,862 \textit{claim–chart–knowledge graph (KG)} triplets, each accompanied by relevant \textit{metadata} (e.g., chart caption, chart type). Claims are systematically constructed based on 2,896 expert-curated scientific charts and annotated with one of three labels: \textit{support}, \textit{refute}, or \textit{not enough information (NEI)}. A key innovation of the dataset is the inclusion of chart-specific knowledge graphs that provide structured, interpretable explanations for fact verification. These KGs capture core scientific information aligned with the claims, such as quantities, trends, spatial and temporal contexts, and causal relationships—enabling explicit multi-hop reasoning. To construct \name, we launched a large-scale project on the citizen science platform Zooniverse\footnote{\url{https://www.zooniverse.org/}}, which ensured scientifically literate annotations. Each chart was independently annotated by six contributors, and the resulting claims were reviewed and verified by two domain experts to ensure correctness and quality.

We utilize \name as a diagnostic benchmark to evaluate the zero-shot and in-context learning capability across a varied range of state-of-the-art models, including open- and closed-source language models, and chart-based vision-language models.
Comprehensive experiments reveal that all models struggle with verifying claims over scientific charts when statistical reasoning is necessary.
Furthermore,
the integration of a chart-specific knowledge graph proves beneficial when models are provided with both scientific charts and supplementary KG data.
In addition, while models generate semantically plausible explanatory triplets, they typically fail to produce properly canonicalized outputs.
These findings underscore the unique challenges posed by \name and highlight the need for further advances in models capable of statistical reasoning, structured explanation generation, and deep scientific understanding from visual evidence.

\begin{table*}[t]
\centering
\small
\setlength{\tabcolsep}{5pt}
\renewcommand{\arraystretch}{1.1}
\begin{tabular}{@{}lccrl@{}}
\toprule
\textbf{Dataset} & \textbf{Modality} & \textbf{Domain} & \textbf{\#Claims} & \textbf{Source} \\
\midrule
SciFact \citep{wadden-2020-scifact} & Text & Biomedical & 1.4K & Medical literature \\
Climate-FEVER \citep{diggelmann2021climatefeverdatasetverificationrealworld} & Text & Climate & 1.5K & Wikipedia \\
HealthVer \citep{healthver-2021-evidence-based} & Text & Health & 14K & News \\
COVID-Fact \citep{saakyan2021covidfact} & Text & COVID-19 & 4.1K & News \\
CoVERT \citep{2022-covert} & Table & COVID-19 & 10K & Social media \\
SciTab \citep{lu-2023-scitab} & Table & CS & 1.2K & Scientific papers \\
\midrule
\textbf{\name} & \textbf{Chart} & \textbf{Climate} & \textbf{49.8K} & \textbf{Expert-curated scientific charts} \\
\bottomrule
\end{tabular}
\caption{Comparison of scientific fact-checking datasets by modality, domain, claim volume, and source. \textbf{\name} is the first chart-based benchmark at this scale, grounded in real expert-curated scientific charts.}
\label{tab:comparison}
\end{table*}

\section{Related Work}
\paragraph{Scientific Fact-checking Benchmarks.}
Several benchmarks have been proposed to advance automated scientific fact-checking, primarily focusing on textual evidence (see Table \ref{tab:comparison}). SciFact \citep{wadden-2020-scifact} introduced claim verification against biomedical research abstracts, while Climate-FEVER \citep{diggelmann2021climatefeverdatasetverificationrealworld} extended claim verification to the climate domain using Wikipedia articles. Other datasets, such as HealthVer \citep{healthver-2021-evidence-based} and COVID-Fact \citep{saakyan2021covidfact}, collected claims from health news and pandemic-related sources, respectively. More recently, CoVERT \citep{2022-covert} and SciTab \citep{lu-2023-scitab} shifted toward structured evidence using tables from social media and scientific papers. However, these benchmarks largely target shallow reasoning tasks, often allowing claims to be verified through direct evidence matching rather than deeper inferential processes. Moreover, they rely exclusively on textual or tabular evidence and overlook scientific charts, which are central to communicating empirical findings in scientific domains. In contrast, \name introduces a large-scale, expert-verified benchmark grounded in high-quality scientific charts, requiring statistical reasoning for claim verification and aiming to more closely reflect real-world scientific fact-checking scenarios.
 
\paragraph{Fact-checking over Structured and Visual Data.}

Beyond textual evidence, fact-checking over structured formats such as tables and visualizations has attracted growing attention. TabFact \citep{chen-2020-tabfact} introduced a benchmark for fact verification against Wikipedia tables, while FEVEROUS \citep{aly2021feverousfactextractionverification} extended claim verification to semi-structured tables. Models such as TAPAS \citep{Herzigtapas_2020} and DePlot \citep{liu-etal-2023-deplot} enable direct reasoning over tabular data by treating tables as inputs to pretrained language models. In parallel, chart understanding has emerged as a distinct challenge, with datasets like PlotQA \citep{methani2020plotqareasoningscientificplots}, ChartQA \citep{masry2022chartqabenchmarkquestionanswering}, and ChartBench \citep{xu2024chartbenchbenchmarkcomplexvisual} focusing on data extraction or question answering over charts. However, these efforts typically rely on synthetic charts and frame the task narrowly, limiting their relevance for real-world fact-checking~\citep{GuoSV22}. ChartCheck \citep{akhtar2024chartcheckexplainablefactcheckingrealworld} is a more recent dataset that targets fact-checking over Wikimedia charts, but its reliance on non-curated, relatively simple visualizations—mostly line and bar graphs—and its focus on shallow observational claims restricts its depth and utility. In contrast, \name introduces high-quality, expert-curated scientific charts exhibiting greater structural and semantic complexity, and frames fact-checking as a task requiring statistical reasoning over visualized data, offering a significantly more realistic and challenging benchmark for scientific verification.

\paragraph{Statistical Reasoning in NLP.}
Statistical reasoning refers to the process of interpreting, analyzing, and drawing inferences from quantitative data—often involving trends, comparisons, variability, and uncertainty—to reach logically sound conclusions \citep{StatisticalReasoning}. Unlike general reasoning, which may rely on commonsense or world knowledge, statistical reasoning demands precise, data-grounded inference directly from observed evidence. This capability is particularly critical in scientific fact-checking, where verifying claims derived from charts requires understanding and interpreting complex quantitative patterns. While reasoning tasks have been extensively studied in NLP, existing benchmarks rarely require statistical reasoning over charts; most focus on discrete, categorical reasoning \citep{pan-etal-2023-fact, akhtar2023readingreasoningchartimages, glockner-etal-2024-ambifc} rather than interpreting continuous data distributions. Techniques such as few-shot prompting \citep{brown2020languagemodelsfewshotlearners} have shown promise in improving performance on symbolic and arithmetic reasoning tasks, but our experiments demonstrate that few-shot prompting yields minimal gains on \name—underscoring the unique challenges posed by statistical reasoning over scientific charts.

\section{\name: Dataset Construction}
\subsection{Annotation}
We manually selected 2,896 diverse scientific charts from six respected open-domain climate sources, each accompanied by metadata\footnote{\url{https://www.noaa.gov/}, \\ \url{https://www.metoffice.gov.uk/}, \\
\url{https://www.copernicus.eu/},  \\
\url{https://earthobservatory.nasa.gov/}, \\
\url{https://www.climate.gov/}, \\
\url{https://climatereanalyzer.org/}}. These charts—spanning topics such as temperature anomalies, CO\textsubscript{2} concentrations, precipitation trends, and sea level rise—were used to design a three-task annotation project on Zooniverse, a well-established and influential citizen science platform \citep{fortson2011galaxyzoomorphologicalclassification, Simpson2014}. We provided annotators with a comprehensive field guide, golden samples labeled by the authors for pre-annotation training, and a live discussion board to assist with challenging cases during the annotation process (see Appendix \ref{app:annotation}). Each chart was independently annotated by six contributors.

\subsubsection{Chart Type Annotation}
In the first task, annotators were asked to identify the chart type by selecting from a set of predefined categories: line graph, pie chart, scatter plot, geographic map, or other (see Figure \ref{fig:distribution}). Given the complexity of many scientific charts, such as those with overlapping modalities or multiple subplots (see Figure~\ref{fig:sample}), annotators were permitted to select multiple chart types for a single instance. This task aimed to categorize chart forms to support downstream analysis and model conditioning.

\subsubsection{Caption Annotation}
In the second task, annotators were instructed to write or revise the caption associated with each chart, ensuring clarity, accuracy, and conciseness in describing its content. In addition, annotators were asked to compose at least one true claim per chart that required statistical reasoning and was directly verifiable from the visualized data. We applied an automated preprocessing step to filter out incomplete or overly short claims (fewer than 10 words). The remaining claims were manually validated by two domain experts according to two criteria: (i) factual correctness independent of external context, and (ii) direct verifiability using only the information presented in the chart. Claims that met both criteria were labeled as “keep”; those that did not were discarded.

To generate refuted claims, we employed GPT-4o~\citep{gpt4o}, prompting it to apply common data fallacy strategies including trend modification, exaggeration, and metric swaps~\citep{akhtar2024chartcheckexplainablefactcheckingrealworld,xu2024chartbenchbenchmarkcomplexvisual}. To ensure each generated claim was semantically contradictory to the original, we filtered 20,148 candidates from 23,190 generated refuted claims
using DeBERTa-Large-MNLI~\citep{laurer2022debertamnli}. Outputs passing this stage were then reviewed by domain experts to verify grammaticality and falsifiability with respect to the associated chart.

For NEI (Not Enough Information) claims, we employed conceptual generalization~\citep{DrchalPipelineAD}, transforming specific factual details into broader or unverifiable language (e.g., “Florida” → “a coastal region”). We combined 200 manually authored NEI examples with GPT-4o-generated variants, additionally prompting entity replacements (e.g., “average” → “maximum” anomaly) to increase linguistic and semantic diversity. All NEI claims were independently verified by two domain experts to ensure that they were plausible yet unverifiable based on the chart. See Table~\ref{tab:NEIandrefute} for examples of refuted and NEI claims.

\subsubsection{Knowledge Graph-Based Explanation}

We propose a method for generating structured explanations in chart-based fact-checking by constructing chart-specific knowledge graphs (KGs) composed of canonicalized $(h, r, t)$ triplets. In contrast to prior work that applies LLMs to general-purpose knowledge graph construction~\citep{CodeKGC,evaluatingchatgptsinformationextraction}, we use a multimodal LLM (GPT-4o~\citep{gpt4o}) to extract factual triplets.

The pipeline begins by parsing each chart and its caption, followed by aggregating all supported claims to construct a unified chart summary. GPT-4o is then prompted with this context—chart image, metadata, and summary—under a loosely defined schema to extract factual triplets that reflect the chart's content. To reduce ambiguity and improve consistency, we apply a self-canonicalization stage inspired by the Extract, Define, Canonicalize (EDC) framework~\citep{zhang2024extractdefinecanonicalizellmbased}, which standardizes the representation of entities and relations across triplets.

These chart-derived triplets serve as structured and interpretable explanations that support fact-checking decisions. For schema details and representative examples, see Appendix~\ref{sec:appendix-kge}.

\begin{table}[t]
\centering
\small
\begin{tabular}{@{}lr@{}}
\toprule
\textbf{Statistic} & \textbf{Value} \\
\midrule
Supported claims       & 15,100 \\
NEI claims             & 15,258 \\
Refuted claims         & 19,504 \\
\textbf{Total claims}         & \textbf{49,862} \\
\midrule
charts         & 2,896 \\
\midrule
Avg. tokens per claim  & 19.0 \\
Avg. claims per chart  & 17.2 \\
\bottomrule
\end{tabular}
\caption{Dataset statistics for \name.
NEI stands for \textit{Not Enough Information}.}
\label{tab:claim_stats}
\end{table}

\begin{table}[t]
\centering
\small
\begin{tabular}{lc}
\toprule
\textbf{Annotation Task} & \textbf{Randolph’s Kappa} \\ 
\midrule
Chart Type Annotation & 82.9 \\
Caption Annotation & 68.3 \\
Claim Generation & 76.5 \\
\bottomrule
\end{tabular}
\caption{Randolph’s Kappa values for IAA across tasks.}
\label{tab:iaa_results}
\end{table}

\subsection{Dataset Analysis}
\subsubsection{Dataset Statistics}
\name comprises a total of 49,862 claims labeled as \textit{support}, \textit{refute}, or \textit{not enough information (NEI)} against 2,896 expert-curated charts from the Climate field. The statistics of our \textsc{ClimateViz} are shown in Table~\ref{tab:claim_stats}.

We computed inter-annotator agreement scores using Randolph’s Kappa \citep{randolph2005free} across the three annotation tasks. For the first and second tasks, agreement was measured among six annotators per chart, while for the third task, agreement was calculated between two domain experts responsible for validating the final set of claims. The resulting scores (see Table~\ref{tab:iaa_results}) indicate substantial agreement across all tasks \cite{Landis1977TheMO}.
\begin{figure}[htbp]
    \centering
    \includegraphics[width=0.8\linewidth]{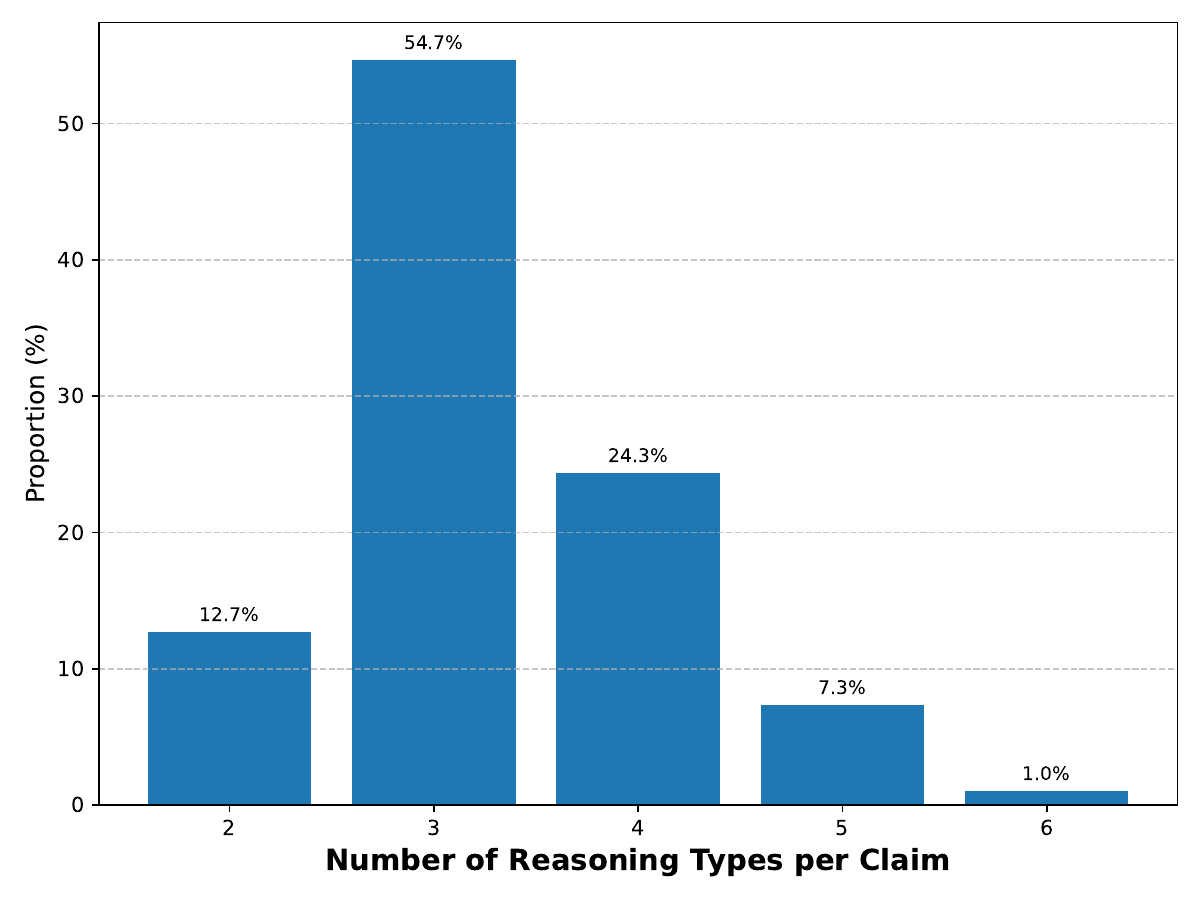} 
    \caption{Distribution of statistical reasoning complexity in \name{} claims.}
    \label{fig:statistical_reasoning}
\end{figure}
\subsubsection{Statistical Reasoning in \name}
We randomly sampled a balanced subset of 300 claims from the \name dataset, covering a diverse range of chart types. Each claim was manually annotated by an author with the types of statistical reasoning required for its verification, following the taxonomy defined in Table~\ref{tab:reasoning_types}.
We observe that temporal comparison, value extraction, and anomaly detection are the most prevalent reasoning types in our dataset.

We further analyze the complexity of claims by examining the number of statistical reasoning types required per instance (Figure~\ref{fig:statistical_reasoning}). Our analysis reveals that multi-hop reasoning is prevalent: a majority of claims (79.0\%) require three or four distinct types of statistical reasoning. This highlights the inherently compositional nature of scientific fact verification in \name.

This deeper analysis demonstrates that \name is not merely a collection of simple lookup tasks. Rather, it challenges models to perform compositional statistical inference—often across varying temporal scales, spatial contexts, and measurement units. As such, \name serves as a rigorous benchmark for evaluating a model’s ability to perform complex, multi-faceted statistical reasoning over scientific visual evidence.

\begin{table*}[t]
\centering
\scriptsize
\renewcommand{\arraystretch}{1.05}
\begin{tabular}{p{2.7cm} p{1.0cm} p{3.8cm} p{4.8cm}}
\toprule
\textbf{Statistical Reasoning} & \textbf{Prop. (\%)} & \textbf{Definition} & \textbf{Example} \\
\midrule
\textbf{Temporal Comparison} & 75.7\% & Compare \textcolor{orange}{values} across \textcolor{blue}{time points}. & temperature in \textcolor{blue}{2020} higher than \textcolor{blue}{2010} \\
\addlinespace[0.5em]
\textbf{Value Extraction} & 63.0\% & Read exact \textcolor{orange}{values} from charts. & CO\textsubscript{2} level was \textcolor{orange}{412 ppm} in 2020 \\
\addlinespace[0.5em]
\textbf{Anomaly Detection} & 52.3\% & Spot unexpected \textcolor{red}{patterns} or \textcolor{orange}{values}. & a \textcolor{red}{sudden spike} in temperature \\
\addlinespace[0.5em]
\textbf{Temporal Aggregation} & 49.3\% & Summarize data over \textcolor{blue}{periods}. & average rainfall over \textcolor{orange}{a decade} \\
\addlinespace[0.5em]
\textbf{Spatial Comparison} & 35.7\% & Compare across different \textcolor{blue}{regions}. & \textcolor{blue}{England} warmer than \textcolor{blue}{Scotland} in July \\
\addlinespace[0.5em]
\textbf{Trend Detection} & 26.3\% & Identify \textcolor{red}{rising or falling trends}. & CO\textsubscript{2} emissions \textcolor{orange}{rising} over time \\
\addlinespace[0.5em]
\textbf{Unit Interpretation} & 14.0\% & Understand and convert \textcolor{purple}{units}. & \textcolor{purple}{mm} of rain converted to \textcolor{purple}{inches} \\
\addlinespace[0.5em]
\textbf{Uncertainty} & 13.0\% & Interpret \textcolor{red}{variability} or \textcolor{red}{error bars}. & temperature estimate: \textcolor{red}{$20^\circ\pm1^\circ$C} \\
\bottomrule
\end{tabular}
\caption{Statistical reasoning types required in \name claims. ``Prop. (\%)'' denotes the proportion of sampled claims.}
\label{tab:reasoning_types}
\end{table*}

\section{Experimental Setup}
\subsection{Task Settings}
We define two input settings for the chart-based fact-checking task:

\begin{itemize}[leftmargin=*, topsep=0pt, itemsep=1pt]
    \item \textbf{Chart + Text (CT):} The model $\mathcal{M}$ receives a chart  $\mathcal{C}_{\text{chart}}$, an associated caption $\mathcal{T}_{\text{caption}}$, and a claim $\mathcal{T}_{\text{claim}}$, and predicts a fact-checking label $\mathcal{Y} \in \{\text{support, refute, NEI}\}$.
    
    \item \textbf{Chart + Table + Text (CTT):} We apply a chart-to-table conversion model, DePlot~\citep{deplot}, to extract a structured table $\mathcal{T}_{\text{table}}$ from $\mathcal{C}_{\text{chart}}$. The model $\mathcal{M}$ then receives $(\mathcal{C}_{\text{chart}}, \mathcal{T}_{\text{table}}, \mathcal{T}_{\text{caption}}, \mathcal{T}_{\text{claim}})$ as input and predicts $\mathcal{Y}$.
\end{itemize}

We further consider two output settings:

\begin{itemize}[leftmargin=*, topsep=0pt, itemsep=1pt]
    \item \textbf{Label-Only Output:} The model outputs only the fact-checking label $\mathcal{Y}$:
    \[
    \mathcal{F}(\text{inputs}) \rightarrow \mathcal{Y}.
    \]
    
    \item \textbf{Explanation-Augmented Output:} The model outputs both a set of structured explanatory triplets $\mathcal{E}$ and the final label $\mathcal{Y}$:
    \[
    \mathcal{F}(\text{inputs}) \rightarrow (\mathcal{E}, \mathcal{Y}).
    \]
\end{itemize}
We evaluate label classification using accuracy and macro F$_1$.
 We further evaluate generated triplets using BLEU \citep{papineni-etal-2002-bleu}, METEOR \citep{banerjee-lavie-2005-meteor}, ROUGE-L 
 \citep{lin-2004-rouge} and BERTScore \citep{zhang2019bertscore}.

To validate the use of DePlot-generated tables, we conducted a manual evaluation of 50 chart-to-table conversions, sampled across five chart types. Each output was assessed according to three criteria: fidelity to the original chart, omission of relevant information, and presence of hallucinated content. Our analysis (see Appendix~\ref{sec:manual-chart2table}) shows that DePlot produces high-fidelity tables for line and bar charts, and reasonably accurate tables for pie charts, scatter plots, and maps. These results support the reliability of $\mathcal{T}_{\text{table}}$ as a structured input in our experimental settings.

\subsection{Baselines}

We evaluate a suite of state-of-the-art multimodal models across multiple configurations.

\paragraph{Open-source models.}
We include three publicly available vision-language models: LLaMA-4-Maverick-400B~\citep{llama4}, InternVL-2.5-78B~\citep{internvl25}, and Qwen2.5-VL-72B~\citep{bai2025qwen25vltechnicalreport}, evaluated under both zero-shot and few-shot settings.

\paragraph{Closed-source models.}
We evaluate three proprietary multimodal large language models (MLLMs): o3~\citep{o3model}, GPT-4o~\citep{gpt4o}, and Gemini 2.5~\citep{gemini2.5}, under both zero-shot and few-shot settings.

\paragraph{Chart-specific vision-language models.}
We include several variants of Matcha~\citep{liu2023matchaenhancingvisuallanguage}, an open-source model designed specifically for chart understanding. In particular, we evaluate two off-the-shelf variants: Matcha-ChartQA, pre-trained on the ChartQA benchmark~\citep{masry2022chartqa}, and Matcha-PlotQA, trained on the PlotQA dataset~\citep{methani2020plotqa}, targeting chart question answering and plot comprehension, respectively. Additionally, we fine-tune the base Matcha model on the \name training and development sets for the fact-checking task.

\paragraph{Human performance.}
To establish an upper bound for model performance, we include a human evaluation baseline. We randomly sample 150 examples for each setting from the \name benchmark. Each example is annotated by a human with expertise in both climate science and natural language processing, using the same input modalities as the corresponding model configuration.

\paragraph{Evaluation protocol.}
The \name dataset is split into training (70\%), development (10\%), and test (20\%) subsets. All models are evaluated on the same held-out test set to ensure fair and consistent comparison across model types and input settings.

\begin{table*}[t]
\centering
\scriptsize
\setlength{\tabcolsep}{3.5pt}
\renewcommand{\arraystretch}{1.05}
\begin{tabular}{p{2.5cm}l l cccc cccc}
\toprule
\textbf{Category} & \textbf{Model} & \textbf{Setting} & 
\multicolumn{4}{c}{\textbf{CT}} & 
\multicolumn{4}{c}{\textbf{CTT}} \\
\cmidrule(lr){4-7} \cmidrule(lr){8-11}
& & & \textbf{Acc-L} & \textbf{F1-L} & \textbf{Acc-E} & \textbf{F1-E} & \textbf{Acc-L} & \textbf{F1-L} & \textbf{Acc-E} & \textbf{F1-E} \\
\midrule
\multirow{6}{2.5cm}{\textbf{Closed-source}}
& o3 & Zero-shot & 59.3 & 58.9 & \textbf{84.6} & \textbf{83.1} & 64.0 & 63.3 & 68.9 & 67.8 \\
& o3 & Few-shot & 61.3 \up & 61.0 \up & \textbf{67.5} \down & \textbf{67.0} \down & 65.5 \up & 64.9 \up & 65.4 \down & 64.5 \down \\
& GPT-4o & Zero-shot & 67.8 & 67.5 & \textbf{68.1} & \textbf{68.2} & 64.3 & 64.0 & 60.2 & 59.2 \\
& GPT-4o & Few-shot & 63.3 \down & 59.5 \down & 64.3 \down & 64.9 \down & \textbf{68.3} \up & \textbf{67.9} \up & 62.8 \up & 61.8 \up \\
& Gemini 2.5 & Zero-shot & \textbf{76.2} & \textbf{75.9} & 73.2 & 71.2 & 57.6 & 57.0 & \textbf{85.7} & 57.3 \\
& Gemini 2.5 & Few-shot & 57.4 \down & 53.8 \down & \textbf{73.3} \up & \textbf{73.9} \up & 56.6 \down & 56.2 \down & 70.4 \down & 70.3 \up \\
\midrule
\multirow{6}{2.5cm}{\textbf{Open-source}}
& LLaMA-4-Maverick-17B & Zero-shot & 39.4 & 29.7 & 47.4 & 43.6 & 47.2 & 45.3 & 52.5 & 49.4 \\
& LLaMA-4-Maverick-17B & Few-shot & 54.5 \up & 51.3 \up & 37.8 \down & 29.7 \down & \textbf{79.4} \up & \textbf{76.9} \up & 57.9 \up & 53.0 \up \\
& InternVL 2.5-78B & Zero-shot & 65.8 & 65.7 & 54.6 & 50.4 & 61.3 & 59.8 & 63.3 & 60.9 \\
& InternVL 2.5-78B & Few-shot & 61.3 \down & 61.4 \down & 63.8 \up & 62.5 \up & \textbf{77.8} \up & \textbf{75.5} \up & 76.4 \up & 73.2 \up \\
& Qwen 2.5-VL-72B & Zero-shot & \textbf{68.3} & \textbf{68.3} & 54.3 & 53.8 & 60.8 & 57.9 & 54.3 & 47.7 \\
& Qwen 2.5-VL-72B & Few-shot & 67.3 \down & 68.0 \same & 65.8 \up & 64.3 \up & \textbf{77.8} \up & \textbf{75.3} \up & 72.1 \up & 70.8 \up \\
\midrule
\multirow{4}{2.5cm}{\textbf{Chart-specific}}
& Matcha-ChartQA & Zero-shot & 34.6 & 33.2 & -- & -- & 31.3 & 30.2 & -- & -- \\
& Matcha-PlotQA-V1 & Zero-shot & 21.3 & 21.7 & -- & -- & 24.5 & 22.4 & -- & -- \\
& Matcha-PlotQA-V2 & Zero-shot & 32.4 & 30.6 & -- & -- & 33.4 & 32.9 & -- & -- \\
& Matcha-\name & Fine-tuned & \textbf{51.2} & \textbf{50.7} & -- & -- & \textbf{50.4} & \textbf{48.6} & -- & -- \\
\midrule
\multicolumn{3}{l}{\textbf{Human Performance}} & \textbf{89.3} & \textbf{89.3} & -- & -- & \textbf{92.7} & \textbf{88.6} & -- & -- \\
\bottomrule
\end{tabular}
\caption{
Accuracy and Macro-F1 scores (\%) on the \textbf{\name} fact-checking benchmark across two input settings and two output formats.
\textbf{CT} (Chart+Text): chart image + caption + claim; 
\textbf{CTT} (Chart+Table+Text): chart image + extracted table + caption + claim.
\textbf{Acc-L}/\textbf{F1-L}: label-only output; 
\textbf{Acc-E}/\textbf{F1-E}: explanation-augmented output.
\textbf{Bold} indicates the best score per column. 
$\up$ / $\down$ / $\same$ indicate intra-model differences.
}
\label{tab:fc}
\end{table*}

\section{Results}

\begin{table*}[t]
\centering
\scriptsize
\setlength{\tabcolsep}{3.5pt}
\renewcommand{\arraystretch}{1.05}
\begin{tabular}{@{}l cc cc cc cc cc cc@{}}
\toprule
\multirow{2}{*}{\textbf{Model}} & \multicolumn{4}{c}{\textbf{CT}} & \multicolumn{4}{c}{\textbf{CTT}} \\
\cmidrule(lr){2-5} \cmidrule(lr){6-9}
& \textbf{BLEU} & \textbf{METEOR} & \textbf{ROUGE-L} & \textbf{BERTScore}
& \textbf{BLEU} & \textbf{METEOR} & \textbf{ROUGE-L} & \textbf{BERTScore} \\
\midrule
\textbf{o3 (ZS)} & 20.2 & 66.0 & 57.3 & 92.6 & 21.8 & 66.2 & 56.3 & 92.6 \\
\textbf{o3 (FS)} & 10.3↓ & 53.8↓ & 43.2↓ & 90.3↓ & 11.3↓ & 52.9↓ & 42.5↓ & 91.6↓ \\
\textbf{GPT-4o (ZS)} & \textbf{48.4} & \textbf{77.2} & \textbf{73.6} & 92.7 & \textbf{46.2} & \textbf{73.4} & \textbf{67.4} & 93.2 \\
\textbf{GPT-4o (FS)} & 13.8↓ & 51.3↓ & 38.0↓ & 87.0↓ & 14.7↓ & 55.4↓ & 43.8↓ & 89.4↓ \\
\textbf{Gemini 2.5 (ZS)} & 37.8 & 68.6 & 61.0 & 90.9 & 34.7 & 72.1 & 65.6 & \textbf{92.7} \\
\textbf{Gemini 2.5 (FS)} & 15.2↓ & 57.6↓ & 50.2↓ & 89.7↓ & 15.6↓ & 58.9↓ & 53.8↓ & 89.9↓ \\
\textbf{LLaMA-4-Maverick-17B (ZS)} & 35.3 & 68.3 & 60.3 & 92.3 & 34.8 & 60.2 & 60.2 & 91.3 \\
\textbf{LLaMA-4-Maverick-17B (FS)} & 13.0↓ & 52.2↓ & 41.5↓ & 90.3↓ & 13.1↓ & 48.8↓ & 38.9↓ & 89.4↓ \\
\textbf{InternVL 2.5-78B (ZS)} & 30.8 & 65.2 & 56.6 & 91.4 & 27.6 & 68.1 & 61.2 & 93.1 \\
\textbf{InternVL 2.5-78B (FS)} & 20.7↓ & 65.2$\same$
 & 53.6↓ & 90.6↓ & 23.9↓ & 67.9↓ & 54.1↓ & 91.2↓ \\
\textbf{Qwen 2.5-VL-72B (ZS)} & 36.8 & 66.2 & 57.5 & 91.9 & 35.1 & 70.5 & 61.8 & \textbf{93.5} \\
\textbf{Qwen 2.5-VL-72B (FS)} & 25.7↓ & 57.3↓ & 45.7↓ & 89.3↓ & 9.6↓ & 39.8↓ & 29.9↓ & 88.3↓ \\
\bottomrule
\end{tabular}
\caption{Explanatory triplet generation results on \name. Models are evaluated in both zero-shot (ZS) and few-shot (FS) settings across CT and CTT inputs.$\up$ / $\down$ / $\same$ indicate intra-model change. Bold indicates the best in each column.}
\label{tab:explanation}
\end{table*}

We present the main findings from our experiments in Tables~\ref{tab:fc} and~\ref{tab:explanation}, which report results for both label classification and explanation generation.

\paragraph{Scientific chart-based fact-checking remains challenging for current models.}
Despite recent advances in multimodal reasoning \citep{wang2024exploringreasoningabilitiesmultimodal,wang2025multimodal} and chart understanding \citep{akhtar2024chartcheckexplainablefactcheckingrealworld, xu2024chartbenchbenchmarkcomplexvisual}, a substantial performance gap remains between models and human annotators. Human evaluators achieve 89.3\% accuracy in the \textbf{Chart + Text (CT)} setting and 92.7\% in the \textbf{Chart + Table + Text (CTT)} setting, outperforming all model variants across both input conditions. These results highlight the continued difficulty of scientific chart-based fact verification and the limitations of current models in capturing nuanced statistical reasoning.
\paragraph{Explanation-augmented output improves closed-source model performance.}
Closed-source models such as o3 and Gemini 2.5 show notable gains when generating structured explanations alongside label predictions. For example, o3 achieves the highest explanation-augmented performance in the CT setting, with 84.6\% accuracy and a macro F1 score of 83.1—outperforming all other models. These results suggest that incorporating intermediate reasoning steps enables closed-source models to better ground their predictions, particularly when interpreting complex scientific visualizations.
\paragraph{CTT setting significantly boosts the performance of open-source models under few-shot prompting.}
All open-source models—including LLaMA-4, InternVL 2.5, and Qwen 2.5—achieve their highest label accuracy and F1 scores in the CTT setting when using few-shot prompting. For example, Qwen 2.5 and InternVL both reach 77.8\% accuracy in the CTT few-shot condition, outperforming their CT counterparts. These results highlight the value of structured tabular inputs and prompt-based adaptation for improving factual reasoning in resource-constrained models.
\paragraph{Few-shot prompting offers limited benefit for scientific fact-checking over charts in the CT setting.}
While in-context learning is widely adopted to improve model performance, its impact on scientific fact-checking over charts is inconsistent. Notably, several closed-source models (e.g., GPT-4o and Gemini 2.5) exhibit degraded performance under few-shot prompting in the CT and label-only setting, particularly in explanation and triplet generation tasks. Open-source models also show only marginal gains except for LLaMA-4-Maverick, indicating that a few-shot prompting alone is insufficient to support complex reasoning over scientific visual data.\paragraph{Fine-tuned Matcha-\name performs best among chart-specific models but still lags behind multimodal LLMs.} Among chart-specific baselines, the fine-tuned Matcha-\name model achieves the highest accuracy (51.2\% in CT, 50.4\% in CTT), outperforming zero-shot variants like Matcha-ChartQA and Matcha-PlotQA. However, its performance remains substantially below that of general-purpose multimodal LLMs. This performance gap suggests that while task-specific fine-tuning improves chart understanding, chart-specialized models still lack the general reasoning capabilities and scalability of large multimodal LLMs.
\paragraph{Explanatory triplet generation.}
Across both \textbf{CT} and \textbf{CTT} settings, few-shot prompting consistently degrades triplet quality for all models. GPT-4o remains the strongest performer overall, achieving the highest scores in BLEU, METEOR, and ROUGE-L. While all models attain consistently high BERTScore values—indicating semantic plausibility—their BLEU scores remain low, suggesting that models often generate logically correct triplets but fail to produce outputs in a standard, canonicalized format.

\section{Broader Implications}
\paragraph{NLP for High-Stakes Domains.}
Despite recent advances, current models fall short of human performance in verifying claims from scientific charts, highlighting the need for NLP systems that are both trustworthy and verifiable in high-stakes domains like science communication and policy.
\paragraph{Multimodal and Spatio-Temporal Reasoning.}
\name goes beyond text and tables, requiring reasoning over visual, spatial, and temporal patterns. Current models struggle with this complexity, especially in statistical interpretation, motivating new architectures that unify multimodal reasoning.
\paragraph{Model Explainability.}
\name supports joint evaluation of predictions and reasoning via explanatory triplets. Explanation-augmented outputs improve accuracy in closed-source models, while the gap between BERTScore and BLEU reveals a need for better canonicalization of semantically correct outputs.

\section{Conclusion}
\label{sec:conclusion}
We introduce \name, the first large-scale benchmark for scientific fact-checking grounded in real-world expert-curated charts. By evaluating a diverse range of state-of-the-art models, we reveal limitations in multimodal factual reasoning, especially when statistical interpretation is required. Our findings demonstrate that current models still lag behind human performance, and that in-context learning alone offers limited gains. However, explanation-augmented outputs show promise in improving model reliability and interpretability. \name establishes a new foundation for building multimodal systems that reason faithfully, communicate transparently, and support scientific decision-making in high-stakes domains.

\section*{Limitations}

While \name introduces a comprehensive benchmark for scientific fact-checking over real-world charts and supports structured explanation through knowledge graphs, our study has several limitations.

First, our experiments focus primarily on in-context learning under zero-shot and few-shot settings. We do not explore more advanced prompting strategies such as chain-of-thought (CoT) prompting \citep{wei2022chain}, tree-of-thought (ToT) reasoning \citep{yao2023tree}, or program-guided verification \citep{pan-etal-2023-fact}, which may further improve performance on compositional and multi-hop reasoning tasks. This restricts our ability to fully characterize model capabilities in structured reasoning scenarios.

Second, we evaluate factuality and explanation quality using predefined structured output formats (triplets), but our automatic metrics (e.g., BLEU, METEOR) may not fully capture factual soundness or semantic coherence of the generated explanations~\citep{SchlichtkrullG023}. Future work could incorporate human evaluations or more targeted reasoning metrics.

Lastly, while the dataset spans a wide range of climate topics and chart types, it is domain-specific. Generalization to other scientific disciplines with different conventions, terminologies, or visual formats remains untested.

\section*{Ethics Statement}
We recognize the importance of ethical considerations in our work. 

All charts included in the ClimateViz benchmark are sourced from publicly accessible, reputable scientific institutions, and no proprietary or confidential data was used. The associated claims were annotated through a large-scale citizen science campaign on Zooniverse, with additional quality control by domain experts. Annotators were fully informed about the research purpose and provided their consent voluntarily. No personally identifiable or sensitive information was collected. 

To mitigate potential misuse, particularly in downstream applications that might involve automated verification in policy or legal contexts, we emphasize that the benchmark is intended solely for academic research. Outputs from models trained or evaluated on ClimateViz should not be used in isolation for critical decision-making.

To support reproducibility and responsible use, we released the ClimateViz dataset, code, and documentation under a permissive research license upon publication. 

\bibliographystyle{acl_natbib}
\bibliography{emnlp2023}

\appendix

\section{Annotation for \name}
\label{app:annotation}

\subsection{Before Annotation: Preparation Phase}
Before starting the annotation process, we conducted extensive preparation to ensure that annotators had the necessary guidance, tools, and understanding of the scientific charts. We began with an internal review involving climate science experts and NLP practitioners. This was crucial to refine the scope of the tasks, establish clear goals, and identify potential challenges in the annotation of complex visual scientific data. 

Then, a beta test was conducted with a small group of experienced annotators who provided early feedback on the clarity and difficulty of the tasks. This helped identify areas where instructions or task complexity needed adjustment. Following the beta test, we gathered feedback through detailed forms, allowing us to iteratively improve the task definitions and annotation interface. 

The finalized workflows and task requirements were then implemented on the Zooniverse platform's dedicated webpage, which served as the main point of interaction for annotators.

\subsection{During Annotation: Annotation Phase}
The annotation phase was designed to facilitate a smooth and productive experience for annotators, equipping them with the resources necessary to accurately interpret and label the charts. The tools used during this phase include:

\subsubsection{Field Guide} A comprehensive field guide was provided to the annotators, covering the different types of data representations commonly found in the charts. This guide includes:

\paragraph{Types of Visuals:} Examples of bar charts, line graphs, pie charts, scatter plots, geographic maps, and others, helping annotators become familiar with each format.

\paragraph{Key Definitions:} Explanations of essential concepts, such as "anomalies" or "trends," that might be important when describing climate-related visuals.

\subsubsection{Instructions} Each task was accompanied by explicit, step-by-step instructions. This was especially important for the third task, which involved summarizing factual information from the charts. Annotators were instructed to focus on objective descriptions, providing factual statements that require statistical reasoning regarding the chart without interpretation or bias.

The following are the intructions shown to the annotators for our three tasks.
\begin{tcolorbox}[colback=lightblue, colframe=blue!50!black, 
                  title=\textbf{Task 1: Write a Clear and Informative Caption for the Scientific Chart},
                  sharp corners=south, 
                  boxrule=0.8pt, 
                  arc=4pt,
                  fonttitle=\bfseries, 
                  enhanced]

Welcome! Your task is to write a straightforward, clear caption that accurately describes the main content of the scientific chart. This caption should help a reader quickly understand what the chart shows, without needing to read all the details.

\textbf{Guidelines for Writing the Caption:}

\begin{itemize}
    \item \textbf{Summarize the Main Information:} Focus on the key message or trend shown in the chart. What is the chart primarily about?
    \item \textbf{Use Straightforward English:} Write in plain, clear language without jargon. Your caption should be understandable even to readers outside the field.
    \item \textbf{Ignore Sources and Logos:} Do not include any references to logos, footnotes, or sources. We assume the charts are from reliable resources.
    \item \textbf{If the Chart is Unclear:} If you cannot determine what the chart shows, type ``NA'' as the caption.
    \item \textbf{If Multiple Messages Appear:} If the chart covers multiple topics, focus on the most important or prominent trend or finding.
\end{itemize}

\textbf{Where to Look for Clues:}
\begin{itemize}
    \item \textbf{Chart Title:} Use the chart title if available, but rewrite it slightly to form a complete, descriptive sentence if necessary.
    \item \textbf{Axes Labels:} Look at the x-axis and y-axis labels to understand what is being measured over what range.
    \item \textbf{Legend and Annotations:} If the chart includes a legend or text annotations, use them to guide your description.
\end{itemize}

\end{tcolorbox}

\begin{tcolorbox}[colback=lightblue, colframe=blue!50!black, 
                  title=\textbf{Task 2: Identify the Data Representation Used in the Chart},
                  sharp corners=south, 
                  boxrule=0.8pt, 
                  arc=4pt,
                  fonttitle=\bfseries, 
                  enhanced]

Your next task is to specify how the data in the chart is represented. Some charts use only one form of representation, while others may use several types together. 

Click on all types of data representation that you observe.

Refer to the Field Guide on the right side for detailed descriptions of various data representations used.

\textbf{Available Options:}
\begin{itemize}
    \item Bar Chart
    \item Line Graph
    \item Pie Chart
    \item Scatter Plot
    \item Geographic Map
    \item Other
\end{itemize}

\end{tcolorbox}

\begin{tcolorbox}[colback=lightblue, colframe=blue!50!black, 
                  title=\textbf{Task 3: Write Claims Using Statistical Reasoning Based on the Scientific Chart},
                  sharp corners=south, 
                  boxrule=0.8pt, 
                  arc=4pt,
                  fonttitle=\bfseries, 
                  enhanced]

Your final task is to carefully study the graphic and write one or more factual claims that use \textbf{statistical reasoning}. Each claim should be based directly on what the chart shows.

Imagine you are \textbf{explaining the information to someone who cannot see the graphic}. Your claims should summarize important quantitative patterns, relationships, or trends, using \textbf{straightforward English} and \textbf{specific details} (such as location, time, measurements, and units).

\textbf{Guidelines for Writing Claims:}
\begin{itemize}
    \item \textbf{Base Claims on the Graphic Only:} Use only the information visible in the graphic. Do not rely on outside knowledge.
    \item \textbf{State Quantitative Information Clearly:} Use numbers, percentages, or comparisons whenever possible.
    \item \textbf{Focus on Statistical Trends and Relationships:}
    \begin{itemize}
        \item Changes over time
        \item Comparisons between groups
        \item Visible correlations
    \end{itemize}
    \item \textbf{Be Specific and Detailed:} Include location, time period, and units.
    \item \textbf{One Sentence per Claim:} Write each claim clearly and concisely.
    \item \textbf{Avoid Vague Statements:} Prefer specific, measurable facts.
    \item \textbf{If the Chart is Ambiguous:} Write ``NA'' if you cannot state any confident claims.
\end{itemize}

\textbf{Examples of Good Claims:}
\begin{itemize}
    \item This line graph shows that the average annual temperature in Paris increased from approximately 12°C in 1970 to 15°C in 2020.
    \item The pie chart indicates that over 60\% of global renewable energy production in 2022 came from solar and wind sources combined.
\end{itemize}

\end{tcolorbox}

\subsubsection{Tutorials} We created interactive tutorials that walked annotators through example charts and tasks. These tutorials emphasized how to identify and describe elements like key data points, trends, or anomalies.
\label{app:Annotation for ClimateViz}
\begin{figure}[htbp]
    \centering
    \includegraphics[width=\linewidth]{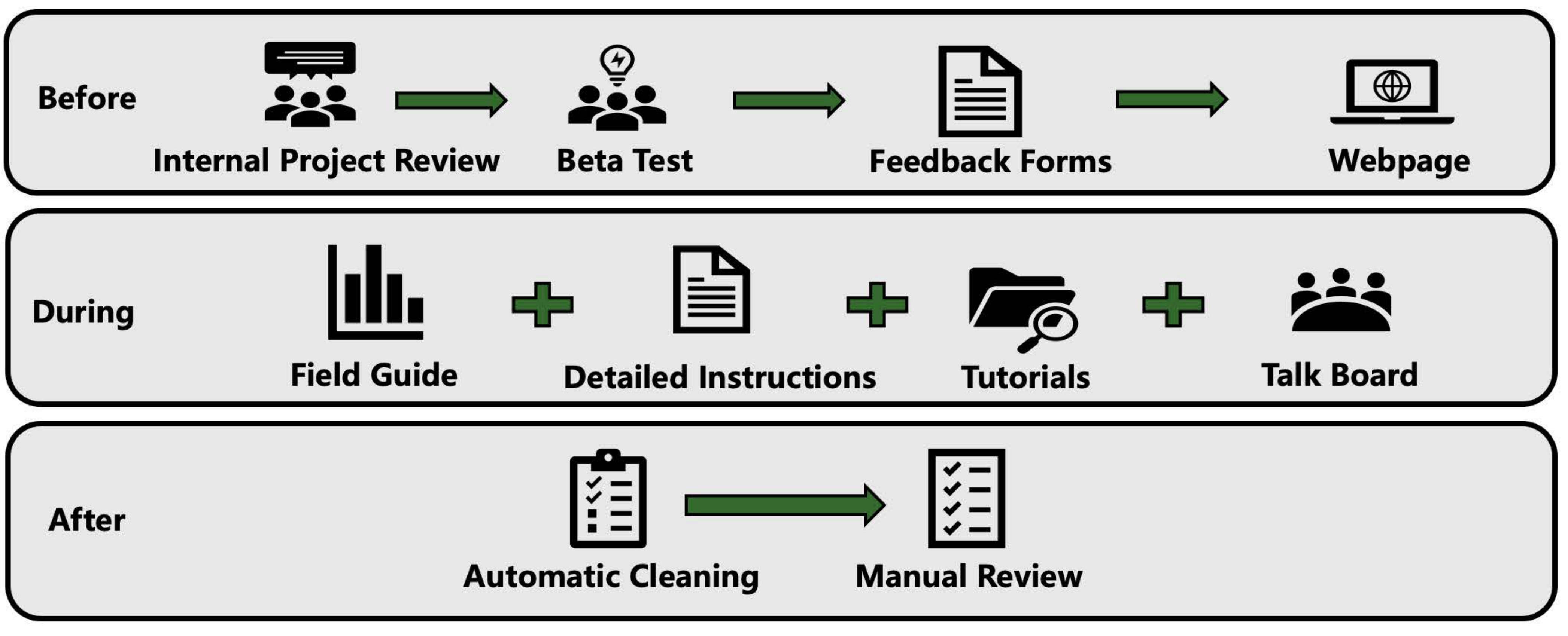} 
    \caption{Quality Control Process: before, during, and after annotation}
    \label{fig:annotation}
\end{figure}
\subsubsection{Talk Board} The Zooniverse platform also included an active "Talk Board" during annotation, where annotators could discuss uncertainties, ask questions, and receive support from both project moderators and their peers. The authors of this paper play an active role in explaining our tasks and discussing how to annotate some particularly complex charts. This collaborative environment was instrumental in resolving ambiguous cases and ensuring consistency across annotations.

\subsection{Post Annotation: Quality Assurance Phase}
Once the annotations were completed, an extensive quality assurance phase was implemented to verify the accuracy and reliability of the collected data.

\subsubsection{Automatic Cleaning} Initially, automated data cleaning scripts were run to detect potential issues such as outlier annotations, incomplete tasks, or incorrect data types. Also, we removed annotations less than 10 words for the "fact" task, with the assumption that they are not informative enough.

\subsubsection{Manual Review} Following the automated cleaning, the data underwent a manual review by domain experts with a PhD degree in climate science and NLP. During this review, we scrutinized the flagged annotations for correctness and consistency. We also went through each claim to make sure it contained the necessary context, which makes it a claim by itself. This dual-step process was critical in catching errors that may have been overlooked by automated methods and ensuring that the dataset retained a high level of reliability.

\section{Statistics for the Scientific Charts}
\label{sec:chart_stats}

The scientific charts used in \name were manually selected from reputable public sources to serve as high-quality, trustworthy visual evidence for fact-checking. All charts were curated to ensure interpretability, sufficient information density, and alignment with key indicators of climate science.

\begin{figure*}[htbp]
    \centering
    \includegraphics[width=\linewidth]{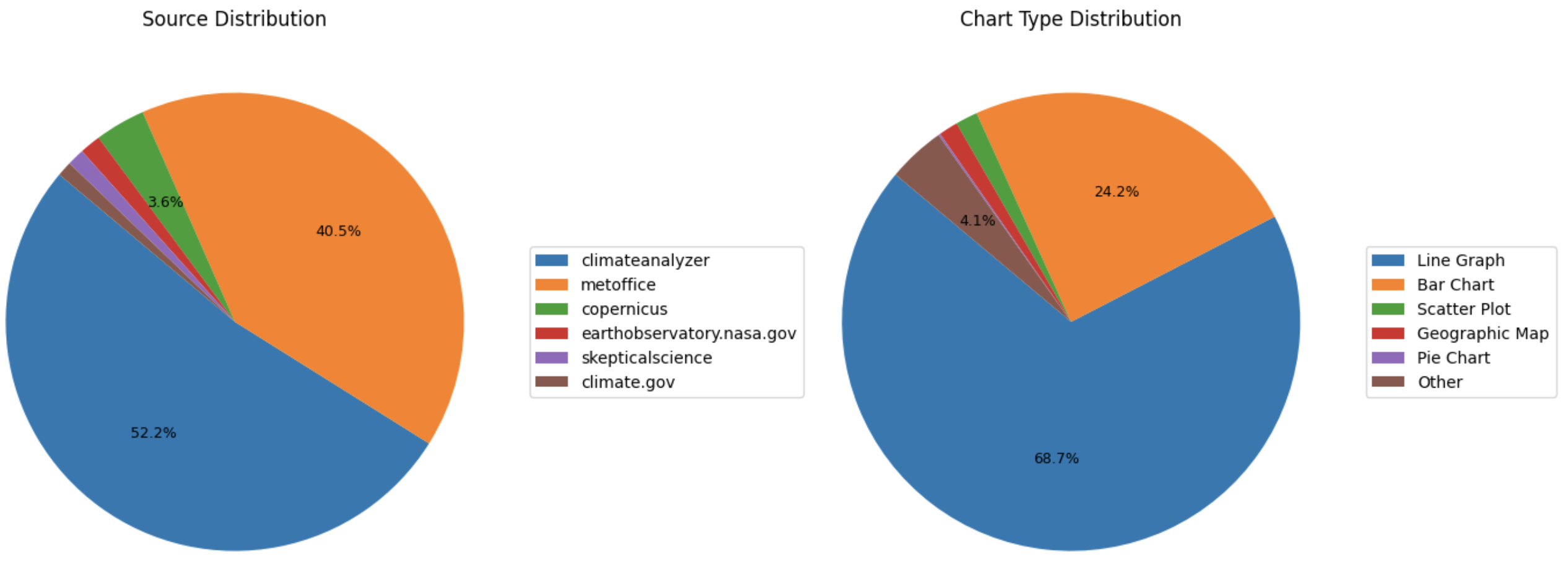} 
    \caption{Distributions of source and type of the charts}
    \label{fig:distribution}
\end{figure*}

Figure~\ref{fig:distribution} presents the distribution of chart sources and chart types. The majority of charts were obtained from two primary sources: climateanalyzer(52.2\%) and the UK Met Office (40.5\%). A smaller proportion of charts come from organizations such as Copernicus, NASA’s Earth Observatory, Skeptical Science, and Climate.gov, each contributing less than 4\%.

In terms of visual representation, line graphs dominate the dataset, comprising 68.7\% of all charts. Bar charts are the second most common (24.2\%), while scatter plots, maps, pie charts, and other types collectively account for the remaining 7.1\%. This reflects the prevalent use of time-series and trend-based data visualization in scientific charts.

By incorporating a wide variety of scientifically valid visualizations from trusted institutions, \name ensures that models are evaluated on realistic and diverse chart-based evidence, closely mirroring the data presentation formats encountered in real-world scientific communication and policymaking.

\label{app:distribution}

\section{Details for Refute and NEI claims}

\subsection{Refuted Claim Generation}
Refuted claims are created by systematically modifying supported claims to introduce factual inaccuracies while maintaining grammatical plausibility. We apply three complementary strategies to generate diverse and realistic refutations:

\paragraph{Trend Modification:} Directional trends in the original claim are reversed to contradict the data. This includes altering keywords such as “increased” to “decreased,” or “rising” to “falling.” These changes invert the implied statistical direction while preserving the overall structure of the sentence.

\paragraph{Exaggeration:} Numerical values and descriptive language are amplified to misrepresent the magnitude of a change. Quantities such as temperature or precipitation are scaled by random multipliers, and qualitative modifiers (e.g., “slight,” “moderate”) are replaced with more extreme terms (e.g., “severe,” “dramatic”).

\paragraph{Metric Swap:} The core metric or variable in the claim is replaced with a similar but distinct one, preserving the sentence form while altering the underlying meaning. For example, “mean maximum temperature” might be swapped with “mean minimum temperature,” or “sunshine duration” replaced by “cloud cover.”

Following generation, we use the DeBERTa-Large-MNLI model~\citep{laurer2022debertamnli} to verify that the modified claim contradicts the original. Each claim pair is scored by the model, which classifies their relationship as entailment, neutral, or contradiction. Only claims labeled as contradiction with high confidence (score > 0.8) are retained.

All accepted refuted claims are then manually reviewed by two domain experts to ensure: (i) the claim is grammatically and semantically well-formed, and (ii) the statement is clearly refuted by the corresponding chart evidence.

\subsection{NEI Claim Generation}
NEI (Not Enough Information) claims are constructed to appear plausible while being unverifiable based on the chart alone. We adopt a multi-step generation strategy combining conceptual generalization, entity replacement, and LLM-based generation:

\paragraph{Conceptual Generalization:} Specific, verifiable references in factual claims are replaced with broader or vaguer terms to obscure direct traceability to the chart. For instance, geographic entities like “Florida” are generalized to “a coastal region,” and temporal references such as “July 2020” are broadened to “a recent summer month.”

\paragraph{Entity Replacement:} Key variables or metrics are substituted with related but unverifiable alternatives. For example, “average temperature anomaly” may be swapped with “maximum temperature anomaly,” or “total precipitation” replaced with “cloud cover.” This ensures the claim remains topically relevant but cannot be definitively supported or refuted by the chart.

\paragraph{LLM-Based Generation:} We manually curated 200 NEI claims and then used GPT-4o~\citep{gpt4o} to generate additional NEI examples by prompting the model with existing NEI instances and instructing it to maintain plausibility while avoiding chart-verifiable details. We included prompts to encourage diversity in language and structure while preserving the overall scientific tone.

All generated NEI claims were filtered to remove overly vague or clearly irrelevant instances. Two domain experts manually validated each claim using the following criteria: (i) the claim must be semantically and grammatically correct, and (ii) it must not be directly classifiable as support or refute based on the chart. Only claims meeting both criteria were included in the final NEI set.

Here are some real examples in the dataset, see Table \ref{tab:NEIandrefute}.

\begin{table*}[t]
\centering
\small
\begin{tabular}{@{}p{5.4cm} p{3.2cm} p{5.4cm}@{}}
\toprule
\textbf{Original Claim} & \textbf{Method} & \textbf{Refuted or NEI Claim} \\
\midrule
\multicolumn{3}{l}{\textbf{Refuted Claims}} \\
\midrule
The mean winter temperature in Wales has shown an upward trend from 1890 to 2020. & Trend Modification & The mean winter temperature in Wales has shown a \textcolor{red}{downward} trend from 1890 to 2020. \\
\addlinespace
The general trend line for sunshine duration in Northern Ireland during spring suggests a slight upward shift over time since 1890. & Exaggeration & The general trend line for sunshine duration in Northern Ireland during spring suggests a \textcolor{red}{significant} upward shift over time since 1890. \\
\addlinespace
Average sunshine duration in August 2021 for England was approximately 186.6 hours. & Metric Swap & Average \textcolor{red}{maximum temperature} in August 2021 for England was approximately 186.6 hours. \\
\midrule
\multicolumn{3}{l}{\textbf{NEI (Not Enough Information) Claims}} \\
\midrule
The average temperature anomaly in April 2015 in Florida was around +3°F. & Conceptual Generalization & The average temperature anomaly in April 2015 in \textcolor{blue}{a coastal region} was around +3°F. \\
\addlinespace
The average temperature anomaly in April 2015 in Florida was around +3°F. & Conceptual Generalization & The average temperature anomaly in \textcolor{blue}{2015} in Florida was around +3°F. \\
\addlinespace
The average temperature anomaly in April 2015 in Florida was around +3°F. & Entity Replacement & The \textcolor{blue}{maximum} temperature anomaly in April 2015 in Florida was around +3°F. \\
\bottomrule
\end{tabular}
\caption{Examples of generating \textbf{refuted} and \textbf{NEI (not enough information)} claims from original climate statements using different perturbation strategies. Color highlights the modified elements (\textcolor{red}{red} for refuted, \textcolor{blue}{blue} for NEI).}
\label{tab:NEIandrefute}
\end{table*}

\section{Knowledge Graph-Based Explanation}
\label{sec:appendix-kge}

To support interpretable and structured scientific fact verification, we construct a knowledge graph (KG) for each chart in \name. These graphs consist of canonicalized triplets of the form $(h, r, t)$—representing factual assertions extracted from chart content. This appendix details our triplet-centric schema, the construction pipeline, and the canonicalization process.

\subsection{Triplet-Aligned Schema}

Each KG is structured as a set of atomic triplets $(h, r, t)$, where:
\begin{itemize}[leftmargin=*, noitemsep]
    \item \textbf{head ($h$):} a scientific entity (e.g., ``Greenland ice sheet''),
    \item \textbf{relation ($r$):} a semantic predicate (e.g., ``contributes to'', ``amount'', ``experienced''), and
    \item \textbf{tail ($t$):} a value, indicator, or another entity (e.g., ``sea level rise'', ``3900 Gt'').
\end{itemize}

To preserve key scientific details, each triplet is accompanied by a \texttt{metadata} object that captures contextual qualifiers such as time period, unit, trend, and uncertainty. This separation enables clear logical reasoning while preserving fidelity to the original chart.

\vspace{0.5em}
\noindent Full triplets based on Figure~\ref{fig:sample} are shown below:
\begin{lstlisting}[language=json, basicstyle=\ttfamily\small]
{
  "triplets": [
    {
      "head": "Greenland Ice Sheet",
      "relation": "experienced",
      "tail": "cumulative mass loss",
      "metadata": {
        "head_type": "Region",
        "tail_type": "Indicator",
        "time_range": "1979--2022",
        "temporal_granularity": "yearly"
      }
    },
    {
      "head": "cumulative mass loss",
      "relation": "trend",
      "tail": "decreasing",
      "metadata": {
        "head_type": "Indicator",
        "tail_type": "Trend",
        "time_range": "2000--2020"
      }
    },
    {
      "head": "cumulative mass loss",
      "relation": "contributes to",
      "tail": "sea level rise",
      "metadata": {
        "head_type": "Indicator",
        "tail_type": "Indicator",
        "time_range": "2000--2020"
      }
    },
    {
      "head": "sea level rise",
      "relation": "amount",
      "tail": "14 mm",
      "metadata": {
        "head_type": "Indicator",
        "tail_type": "Physical Measurement",
        "unit": "mm",
        "time_range": "2020",
        "temporal_granularity": "yearly",
        "uncertainty": "1 mm"
      }
    }
  ]
}
\end{lstlisting}

\subsection{KG Construction Pipeline}

We construct triplets automatically using GPT-4o \citep{gpt4o}, using the chart, caption, and the set of supported claims as the chart summary as inputs. We formulate prompts using a lightly constrained schema, instructing the model to extract semantically grounded $(h, r, t)$ triplets with associated metadata fields. 

\subsection{Self-Canonicalization with LLMs}

Following extraction, we canonicalize the surface forms of both entities and relations. Inspired by the Extract, Define, Canonicalize (EDC) framework~\citep{zhang2024extractdefinecanonicalizellmbased}, we prompt the model to define and normalize semantically equivalent terms. The canonicalized form is extracted from chart captions and summaries. For instance:
\begin{itemize}[leftmargin=*, noitemsep]
    \item ``was about'' $\rightarrow$ \texttt{amount}
    \item ``led to'' $\rightarrow$ \texttt{contributes to}
    \item ``Greenland'' $\rightarrow$ \texttt{Greenland Ice Sheet} 
\end{itemize}

This normalization enables consistency across charts and supports downstream evaluation using structured matching.

\subsection{Coverage and Format}

Triplets are generated only for supported claims to ensure factual consistency with the chart evidence. On average, each chart yields 6–8 canonicalized triplets. The final knowledge graphs are stored in structured JSON files, with each entry linked to its corresponding chart ID. These structured explanations serve dual purposes: (i) enhancing model interpretability and (ii) supporting multi-hop reasoning during fact verification.

These triplets capture causal and quantitative relationships essential for verifying scientific claims and provide a structured representation of the underlying chart semantics.

\subsection{Limitations and Future Work}

While the pipeline produces semantically coherent triplets, errors may arise from ambiguous captions or overloaded visual encodings. In future work, we aim to improve schema alignment with external scientific ontologies, introduce confidence scoring per triplet, and extend the pipeline to cover refuted and NEI claims for contrastive reasoning.

\section{Manual Evaluation for Chart-to-Table Conversion}
\label{sec:manual-chart2table}
This design is motivated by recent work showing that supplementing visual inputs with structured tabular representations improves multimodal reasoning abilities. We aim to study whether this trend also holds for the task of chart-based fact-checking.

To ensure the reliability of our chart-to-table conversions using DePlot, we conduct a manual evaluation of a representative subset of charts in the \name dataset. We randomly sample 50 charts, stratified by chart type: 10 each from \textit{line graph}, \textit{bar chart}, \textit{scatter plot}, \textit{map}, and \textit{pie chart}.

Each generated table is evaluated against three criteria:

\paragraph{Fidelity} Does the table faithfully represent all relevant data values from the chart (e.g., axes, legends, numerical values)?
\paragraph{Omission/Misread} Does the table omit or misinterpret any visual content (e.g., missing labels or incorrect numeric entries)?
\paragraph{Hallucination} Does the table introduce spurious values or labels not present in the original chart?

Based on these criteria, we assign each chart-table pair to one of three categories: \textit{Fully Accurate}, \textit{Minor Issues}, or \textit{Major Issues}.

\begin{table*}[t]
\centering
\small
\setlength{\tabcolsep}{10pt}
\renewcommand{\arraystretch}{1.15}
\begin{tabular}{lcccc}
\toprule
\textbf{Chart Type} & \textbf{\# Samples} & \textbf{Fully Accurate} & \textbf{Minor Issues} & \textbf{Major Issues} \\
\midrule
Line Graph    & 10 & 8 & 1 & 1 \\
Bar Chart     & 10 & 7 & 2 & 1 \\
Scatter Plot  & 10 & 5 & 3 & 2 \\
Map           & 10 & 4 & 4 & 2 \\
Pie Chart     & 10 & 5 & 3 & 2 \\
\midrule
\textbf{Total} & \textbf{50} & \textbf{29} & \textbf{13} & \textbf{8} \\
\bottomrule
\end{tabular}
\caption{Manual evaluation of DePlot’s chart-to-table outputs across five chart types. “Fully Accurate” indicates complete table fidelity; “Minor Issues” include small omissions or rounding mismatches; “Major Issues” involve missing core information or hallucinated values.}
\label{tab:manual-deplot}
\end{table*}

We find that DePlot performs reliably on line and bar charts, where the data structure is linear and labeling is clear. It struggles more with pie charts, maps, and scatter plots, often due to overlapping text, spatial encoding, or small font sizes. This evaluation provides a level of trust in DePlot outputs while acknowledging limitations, especially for spatial or complex chart types.

\section{Experiments}
\subsection{Prompt Templates}
\label{sec:appendix-prompts}

This appendix provides the full prompt templates used in our experiments across different settings. Each template reflects the exact structure used to prompt models in zero-shot and few-shot configurations. For few-shot settings, we include two demonstrations per class label (support, refute, and not enough information) to ensure balance.

\subsubsection{Zero-Shot Prompt (CT, Label-Only Output)}

\begin{promptblock}
\textbf{Instruction:} You are a scientific fact-checking assistant. Based on the chart caption and the claim, determine whether the claim is supported by the chart, refuted by the chart, or if there is not enough information. Respond with one of: \texttt{support}, \texttt{refute}, or \texttt{not enough information}.

\textbf{Caption:} Between 2000 and 2020, the Greenland Ice Sheet experienced accelerating mass loss, contributing to sea level rise. \\
\textbf{Claim:} The Greenland Ice Sheet saw stable mass over the period 2000–2020. \\
\textbf{Answer:}
\end{promptblock}

\subsubsection{Few-Shot Prompt (CT, Label-Only Output)}

\textit{Includes two examples per label. The final query appears after all six examples.}

\begin{promptblock}
\textbf{Example 1} \\
\textbf{Caption:} Average CO\textsubscript{2} levels rose from 370 ppm in 2000 to 412 ppm in 2020. \\
\textbf{Claim:} CO\textsubscript{2} levels have increased between 2000 and 2020. \\
\textbf{Answer:} \texttt{support}
\end{promptblock}

\begin{promptblock}
\textbf{Example 2} \\
\textbf{Caption:} In 2022, the UK experienced the highest annual mean temperature on record. \\
\textbf{Claim:} The UK recorded its coldest year in 2022. \\
\textbf{Answer:} \texttt{refute}
\end{promptblock}

\begin{promptblock}
\textbf{Example 3} \\
\textbf{Caption:} The Arctic sea ice extent varied significantly between 1979 and 2020, with notable seasonal fluctuations. \\
\textbf{Claim:} Arctic sea ice was 5 million sq km in 1999. \\
\textbf{Answer:} \texttt{not enough information}
\end{promptblock}

\begin{promptblock}
\textbf{Example 4} \\
\textbf{Caption:} Annual rainfall in Southern England fluctuated with no clear trend over the last 50 years. \\
\textbf{Claim:} Annual rainfall in Southern England decreased significantly since 1970. \\
\textbf{Answer:} \texttt{refute}
\end{promptblock}

\begin{promptblock}
\textbf{Example 5} \\
\textbf{Caption:} Spring temperature anomalies in Scotland increased slightly between 1960 and 2020. \\
\textbf{Claim:} Scotland saw the largest anomaly in 1978. \\
\textbf{Answer:} \texttt{not enough information}
\end{promptblock}

\begin{promptblock}
\textbf{Example 6} \\
\textbf{Caption:} Average summer temperature in Wales increased by 1.2°C from 1980 to 2020. \\
\textbf{Claim:} Summer temperature in Wales has warmed in the past 40 years. \\
\textbf{Answer:} \texttt{support}
\end{promptblock}

\begin{promptblock}
\textbf{Final Query} \\
\textbf{Caption:} [Tcaption] \\
\textbf{Claim:} [Tclaim] \\
\textbf{Answer:}
\end{promptblock}

\subsubsection{Few-Shot Prompt (CTT, Explanation-Augmented Output)}

\textit{Includes two examples per label with structured triplet explanations. The model must generate both reasoning triplets and the final label.}

\begin{promptblock}
\textbf{Example 1} \\
\textbf{Caption:} Annual mean surface temperature in England from 2015 to 2020. \\
\textbf{Table:}
\begin{center}
\begin{tabular}{|c|c|} \hline
\textbf{Year} & \textbf{Temperature (°C)} \\ \hline
2015 & 9.5 \\
2016 & 9.7 \\
2017 & 9.6 \\
2018 & 9.9 \\
2019 & 10.1 \\
2020 & 10.2 \\ \hline
\end{tabular}
\end{center}

\textbf{Claim:} England's mean surface temperature rose steadily from 2015 to 2020. \\
\textbf{Triplets:}
\begin{itemize}
\setlength\itemsep{0.2em}
\item (England, Experienced, Surface Temperature Increase)
\item (Surface Temperature, Trend, Increasing)
\item (Time Period, Range, 2015–2020)
\item (Temperature, Start Year Value, 9.5°C)
\item (Temperature, End Year Value, 10.2°C)
\item (Increase Amount, Computed Difference, 0.7°C)
\end{itemize}
\textbf{Label:} \texttt{support}
\end{promptblock}

\begin{promptblock}
\textbf{Example 2} \\
\textbf{Caption:} Total rainfall in Wales from 2010 to 2015. \\
\textbf{Table:}
\begin{center}
\begin{tabular}{|c|c|} \hline
\textbf{Year} & \textbf{Rainfall (mm)} \\ \hline
2010 & 1400 \\
2011 & 1380 \\
2012 & 1450 \\
2013 & 1390 \\
2014 & 1420 \\
2015 & 1410 \\ \hline
\end{tabular}
\end{center}

\textbf{Claim:} Wales received significantly less rainfall in 2015 compared to earlier years. \\
\textbf{Triplets:}
\begin{itemize}
\setlength\itemsep{0.2em}
\item (Wales, Experienced, Rainfall)
\item (Time Period, Range, 2010–2015)
\item (Rainfall, Value in 2015, 1410 mm)
\item (Rainfall, Mean Value 2010–2014, 1408 mm)
\item (Rainfall in 2015, Comparative Trend, No Significant Decrease)
\item (Rainfall Comparison, Difference from Average, +2 mm)
\end{itemize}
\textbf{Label:} \texttt{refute}
\end{promptblock}

\begin{promptblock}
\textbf{Example 3} \\
\textbf{Caption:} Annual sunshine duration in Scotland between 1995 and 2000. \\
\textbf{Table:}
\begin{center}
\begin{tabular}{|c|c|} \hline
\textbf{Year} & \textbf{Sunshine Hours} \\ \hline
1995 & 1100 \\
1996 & 1080 \\
1997 & 1095 \\
1998 & 1120 \\
1999 & 1090 \\
2000 & 1085 \\ \hline
\end{tabular}
\end{center}

\textbf{Claim:} Scotland had the highest annual sunshine duration on record in 2001. \\
\textbf{Triplets:}
\begin{itemize}
\setlength\itemsep{0.2em}
\item (Scotland, Recorded, Sunshine Duration)
\item (Time Period, Table Coverage, 1995–2000)
\item (Sunshine Duration in 2001, Availability, Missing)
\item (Max Sunshine in Table, Year, 1998)
\item (Assertion Year 2001, Not Covered in Table, True)
\end{itemize}
\textbf{Label:} \texttt{not enough information}
\end{promptblock}

\begin{promptblock}
\textbf{Example 4} \\
\textbf{Caption:} Annual CO\textsubscript{2} concentrations globally from 2010 to 2015. \\
\textbf{Table:}
\begin{center}
\begin{tabular}{|c|c|} \hline
\textbf{Year} & \textbf{CO\textsubscript{2} (ppm)} \\ \hline
2010 & 390.1 \\
2011 & 392.6 \\
2012 & 395.4 \\
2013 & 397.9 \\
2014 & 399.8 \\
2015 & 402.3 \\ \hline
\end{tabular}
\end{center}

\textbf{Claim:} CO\textsubscript{2} levels increased each year from 2010 to 2015. \\
\textbf{Triplets:}
\begin{itemize}
\item (Global Atmosphere, Measured, CO\textsubscript{2})
\item (Time Period, Range, 2010–2015)
\item (CO\textsubscript{2}, Trend, Increasing)
\item (CO\textsubscript{2} in 2010, Value, 390.1 ppm)
\item (CO\textsubscript{2} in 2015, Value, 402.3 ppm)
\end{itemize}
\textbf{Label:} \texttt{support}
\end{promptblock}

\begin{promptblock}
\textbf{Example 5} \\
\textbf{Caption:} Average spring temperatures in Northern Ireland from 2000 to 2005. \\
\textbf{Table:}
\begin{center}
\begin{tabular}{|c|c|} \hline
\textbf{Year} & \textbf{Temperature (°C)} \\ \hline
2000 & 8.2 \\
2001 & 8.3 \\
2002 & 8.5 \\
2003 & 8.7 \\
2004 & 8.9 \\
2005 & 9.0 \\ \hline
\end{tabular}
\end{center}

\textbf{Claim:} Spring temperatures in Northern Ireland gradually increased from 2000 to 2005. \\
\textbf{Triplets:}
\begin{itemize}
\item (Northern Ireland, Experienced, Spring Temperature Increase)
\item (Time Period, Range, 2000–2005)
\item (Spring Temperature, Trend, Increasing)
\item (Spring Temperature in 2000, Value, 8.2°C)
\item (Spring Temperature in 2005, Value, 9.0°C)
\item (Increase Amount, Computed, 0.8°C)
\end{itemize}
\textbf{Label:} \texttt{support}
\end{promptblock}

\begin{promptblock}
\textbf{Example 6} \\
\textbf{Caption:} Monthly average rainfall in Scotland in 2022. \\
\textbf{Table:}
\begin{center}
\begin{tabular}{|c|c|} \hline
\textbf{Month} & \textbf{Rainfall (mm)} \\ \hline
Jan & 120 \\
Feb & 115 \\
Mar & 90 \\
Apr & 85 \\
May & 75 \\
Jun & 65 \\
Jul & 70 \\
Aug & 80 \\
Sep & 95 \\
Oct & 110 \\
Nov & 125 \\
Dec & 130 \\ \hline
\end{tabular}
\end{center}

\textbf{Claim:} Rainfall in Scotland was highest in winter months during 2022. \\
\textbf{Triplets:}
\begin{itemize}
\setlength\itemsep{0.2em}
\item (Scotland, Observed, Monthly Rainfall)
\item (Winter Months, Include, Dec–Feb)
\item (Rainfall, Highest Values, Dec:130mm, Jan:120mm, Nov:125mm)
\item (Winter Rainfall, Compared to, Higher than Summer)
\item (Time Period, Year, 2022)
\item (Rainfall, Seasonal Trend, Peak in Winter)
\end{itemize}
\textbf{Label:} \texttt{support}
\end{promptblock}

\begin{promptblock}
\textbf{Final Query} \\
\textbf{Caption:} [Tcaption] \\
\textbf{Table:} [Ttable] \\
\textbf{Claim:} [Tclaim] \\
\textbf{Triplets:} \\
\textbf{Label:}
\end{promptblock}

\subsection{Table + Text Only Ablation.}
To evaluate the importance of visual input in scientific chart-based fact-checking, we conduct an ablation study by removing the chart image and providing only the structured table (generated via DePlot) along with the chart caption and the claim as model input. This setting, denoted as \textbf{Table + Text}, isolates the contribution of the tabular and textual modalities, allowing us to assess whether models can accurately verify claims without access to the original chart. While this setup preserves key quantitative patterns through table representations, it lacks access to spatial, visual, and stylistic cues embedded in the chart. Our results indicate that performance drops noticeably compared to the full \textbf{Chart + Table + Text} (CTT) setting, highlighting the complementary role of visual features in supporting accurate and interpretable fact verification.
\begin{table*}[t]
\centering
\small
\setlength{\tabcolsep}{5pt}
\renewcommand{\arraystretch}{1.1}
\begin{tabular}{@{}l l c c@{}}
\toprule
\textbf{Model} & \textbf{Setting} & \textbf{Acc-L (TT)} & \textbf{F1-L (TT)} \\
\midrule
\multicolumn{4}{l}{\textbf{Closed-source}} \\
\midrule
o3                        & Zero-shot  & 47.6 & 43.7 \\
                         & Few-shot   & 49.2~($\uparrow$) & 45.1~($\uparrow$) \\
GPT-4o                   & Zero-shot  & 51.5 & 48.0 \\
                         & Few-shot   & 52.9~($\uparrow$) & 49.2~($\uparrow$) \\
Gemini 2.5               & Zero-shot  & \textbf{53.5} & \textbf{53.4} \\
                         & Few-shot   & 52.2~($\downarrow$) & 52.5~($\downarrow$) \\
\midrule
\multicolumn{4}{l}{\textbf{Open-source}} \\
\midrule
LLaMA-4-Maverick-17B     & Zero-shot  & 47.2 & 45.3 \\
                         & Few-shot   & 52.5~($\uparrow$) & 49.7~($\uparrow$) \\
InternVL 2.5-78B         & Zero-shot  & 53.3 & 49.1 \\
                         & Few-shot   & \textbf{55.6}~($\uparrow$) & \textbf{51.0}~($\uparrow$) \\
Qwen 2.5-VL-72B          & Zero-shot  & 52.8 & 48.8 \\
                         & Few-shot   & 54.4~($\uparrow$) & 51.2~($\uparrow$) \\
\bottomrule
\end{tabular}
\caption{Ablation study results for the \textbf{Table + Text (TT)} setting, where models receive only the DePlot-generated table, chart caption, and claim, omitting the chart image. Acc-L and F1-L denote label-only accuracy and macro F1, respectively. Arrows indicate intra-model performance change from zero-shot to few-shot. Bolded values are the best per column.}
\label{tab:ablation-tt}
\end{table*}

Table~\ref{tab:ablation-tt} presents the ablation results under the Table + Text (TT) setting, where the chart image is omitted. Across both open- and closed-source models, we observe that performance declines moderately compared to the full CTT setting, indicating that visual input contributes complementary information beyond structured tabular data. InternVL 2.5 and Qwen 2.5 achieve strong performance, with InternVL reaching the highest accuracy (55.6\%) in the few-shot condition. Interestingly, Gemini 2.5 yields the best zero-shot results (53.5\% accuracy, 53.4\% F1), but performance degrades slightly with few-shot prompting—mirroring the instability seen in other few-shot settings. Among open-source models, all benefit consistently from few-shot prompting, whereas closed-source models exhibit marginal gains or regressions. These results confirm that while structured table representations alone are effective, incorporating chart visuals remains essential for optimal fact verification performance.

\end{document}